\pdfoutput=1
\documentclass{article}

\usepackage{microtype}
\usepackage{graphicx}
\graphicspath{{figures/}}
\usepackage{booktabs}
\usepackage{amsmath,amssymb}
\usepackage{multirow}
\usepackage{multicol}
\usepackage{siunitx}
\usepackage{hyperref}
\usepackage{xcolor}
\usepackage{natbib}
\usepackage{placeins}
\usepackage{caption}
\usepackage{capt-of}
\usepackage{tabularx}
\usepackage{ragged2e}
\usepackage{placeins}
\usepackage{dblfloatfix}

\usepackage[accepted]{style/icml2026}
\makeatletter
\renewcommand{\Notice@String}{}
\def\icmlcorrespondingauthor@text{}

\renewcommand{\printAffiliationsAndNotice}[1]{%
  \global\icml@noticeprintedtrue%
  \stepcounter{@affiliationcounter}%
  {\let\thefootnote\relax\footnotetext{%
      \forloop{@affilnum}{1}{\value{@affilnum} < \value{@affiliationcounter}}{%
        \textsuperscript{\arabic{@affilnum}}%
        \csname @affilname\the@affilnum\endcsname
      }.%
    }%
  }%
}
\makeatother

\newcommand{\dataset}[1]{\textsc{#1}}
\newcommand{\model}[1]{\textsc{#1}}

\newcommand{\HCThreePlus}{\dataset{HC3~PLUS}}

\newcommand{\ba}{\ensuremath{\mathrm{BA}}}

\newcommand{\tpr}{\ensuremath{\mathrm{TPR}}}
\newcommand{\tnr}{\ensuremath{\mathrm{TNR}}}

\newcommand{\yhat}{\ensuremath{\hat{y}}}

\icmltitlerunning{Feature-Augmented Transformers for Robust AI-Text Detection Across Domains and Generators}

\begin{document}

\twocolumn[
\icmltitle{Feature-Augmented Transformers for Robust AI-Text Detection Across Domains and Generators}

\begin{icmlauthorlist}
\icmlauthor{Mohamed Mady}{1,2}
\icmlauthor{Johannes Reschke}{2}
\icmlauthor{Björn Schuller}{1,3}
\end{icmlauthorlist}

\icmlaffiliation{1}{Chair of Health Informatics, Technical University of Munich (TUM), Germany }
\icmlaffiliation{2}{OTH Regensburg, Germany }
\icmlaffiliation{3}{GLAM – Group on Language, Audio \& Music, Imperial College London, United Kingdom}


\begin{abstract}
AI-generated text is produced at scale across diverse domains and heterogeneous generators, making robustness to distribution shift a central requirement for reliable detection. We train transformer-based detectors on \dataset{HC3~PLUS} and adopt a deployment-realistic fixed-threshold protocol: a single decision threshold is calibrated on held-out validation data and kept fixed across all downstream test distributions. This protocol reveals that near-ceiling in-domain performance (up to 99.5\% balanced accuracy) degrades significantly under cross-dataset and generator shift, exposing strong complementary failure modes across backbones (human-preserving vs.\ AI-aggressive). Feature augmentation that fuses handcrafted linguistic signals with transformer representations via a learnable attention module substantially improves transfer. While \model{BERT} and \model{RoBERTa} show complementary weaknesses, our best configuration \model{DeBERTa-v3-base+FeatAttn} yields the most balanced and robust profile, reaching 85.9\% balanced accuracy on the multi-domain, multi-generator \dataset{M4} benchmark (81.3\% human recall, 90.5\% AI recall). Multi-seed experiments (5 seeds) confirm high stability with a macro-average of $83.15 \pm 1.04\%$ on \dataset{M4}. Under the exact same fixed-threshold protocol, our model outperforms strong zero-shot baselines (Fast-DetectGPT, RADAR, Log-Rank) by up to +7.22 points. Category-level ablations further show that readability and vocabulary features contribute most to robustness under shift. Overall, these results demonstrate that feature augmentation and a modern DeBERTa backbone significantly outperform earlier BERT/RoBERTa models, while the fixed-threshold protocol, combined with generator-aware error profiling and explicit feature analysis, provides a more realistic and informative assessment of practical detector robustness.
\end{abstract}
]

\printAffiliationsAndNotice{}

\section{Introduction}
\label{sec:intro}

Frontier large language models (LLMs) can generate fluent and persuasive text across a wide range of styles and tasks, accelerating uptake in education, online platforms, and content creation.
This proliferation increases the practical demand for AI-generated text detection in settings such as academic integrity, content moderation, and dataset curation \citep{Chakraborty24-Position,Mitchell23-DetectGPT,Hans24-Binoculars}.
However, real-world deployment is complicated by distribution shift: detectors are routinely applied to writing styles, domains, and generation pipelines that differ from those seen during training, where strong in-domain scores can be a misleading indicator of reliability.

We study supervised binary AI-text detection under such shift using a deployment-oriented evaluation protocol.
Detectors are trained on \dataset{HC3~PLUS} \citep{Su23-HC3Plus} and assessed without any target-domain tuning.
Instead, each model uses a single decision threshold selected on held-out validation data and then kept fixed when transferring to new domains and generators.
We evaluate robustness across three complementary suites: (i) in-domain assessment on \dataset{HC3~PLUS}, (ii) direct cross-dataset transfer to the multi-domain, multi-generator \dataset{M4} benchmark (English) \citep{Wang24-M4}, and (iii) external testing on \dataset{AI-Text-Detection-Pile} \citep{Artem9k24-AITextDetectionPile}.
For \dataset{M4}, we report domain-level and generator-level results to expose failure patterns that are obscured by a single aggregate score.

Beyond standard transformer classifiers (\model{BERT}, \model{RoBERTa}), we investigate feature-augmented detectors that fuse transformer representations with handcrafted linguistic signals (e.g., lexical diversity, POS patterns, readability and punctuation cues, and language-model-derived statistics).
We also include \model{DeBERTa-v3-base} as an additional backbone.
In our experiments, it yields a more balanced transfer profile; one plausible explanation is that the v3 family is pre-trained with ELECTRA-style replaced-token detection \citep{Clark20-ELECTRA,He21-DeBERTaV3}, which may encourage representations that are less sensitive to superficial cues that vary under rewriting and cross-domain shift.

We further strengthen the evaluation with multi-seed stability analysis (5 seeds), direct comparisons against strong zero-shot baselines (Fast-DetectGPT, RADAR, Log-Rank) under the identical fixed-threshold protocol, and comprehensive feature-category ablations that quantify the contribution of individual linguistic signals to robustness under shift.

Overall, our study yields three takeaways.
First, semantic-invariant evaluation is a high stress test that can expose brittleness masked by near-ceiling performance on the source distribution.
Second, calibrating a single validation threshold and keeping it fixed at test time better reflects deployment constraints and reveals domain- and generator-specific failure modes.
Third, augmenting transformer representations with explicit linguistic features can improve transfer robustness, but it may also sharpen the trade-off between preserving human text and detecting diverse generators.
\FloatBarrier
\section{Related Work}
\label{sec:related}

Prior work on AI-generated text detection spans supervised classifiers trained on labelled human--machine data, zero-shot scoring rules derived from language-model probabilities, and provenance-based approaches such as watermarking, with additional work on streaming/online decision-making.
This contribution focuses on supervised detection, with particular emphasis on evaluation reliability under distribution shift and on reporting practices that remain meaningful when target-domain labels are unavailable.

Benchmarks for human--LLM comparison have enabled rapid progress by providing paired human and model responses at scale. HC3 offers human-written and ChatGPT-generated answers for the same prompts across diverse question categories, supporting systematic evaluation of supervised baselines \citep{Guo23-HC3}.
HC3~PLUS extends this idea by introducing semantic-invariant rewrites (e.g., translation, summarisation, paraphrasing) that preserve the underlying meaning while altering the surface form, substantially increasing detection difficulty \citep{Su23-HC3Plus}.
Such settings better reflect deployment conditions in which generated text may be post-edited or re-expressed before reaching a detector.

Supervised detectors typically fine-tune pre-trained encoders such as \model{BERT} and \model{RoBERTa} for binary classification \citep{Devlin19-BERT,Liu19-RoBERTa}.
Although these models can achieve near-ceiling scores in-domain, multiple studies highlight that performance may degrade under domain shift and generator shift, and that aggregate metrics can conceal brittle behaviour \citep{Chakraborty24-Position}.
This motivates evaluation protocols that explicitly probe robustness across domains, generators, and transformation regimes, together with class-wise reporting to expose asymmetric error patterns.

Zero-shot and model-based detectors aim to reduce reliance on labelled data by exploiting language-model probability structure.
DetectGPT uses the curvature of the log-probability function to separate sampled text from human text without supervised training \citep{Mitchell23-DetectGPT}.
Binoculars constructs a strong zero-shot detector using a pair of related language models and reports competitive performance across multiple generators \citep{Hans24-Binoculars}.
These approaches are complementary to supervised methods, but their robustness can depend on model mismatch and on shifts in the target distribution.

Beyond offline classification, online detection considers sequential decisions as tokens arrive.
Sequential hypothesis testing via betting provides a principled framework, making the trade-off between detection speed and certainty explicit \citep{Chen25-OnlineBetting}.
While the present study is offline, the same deployment pressures arise as generators and target distributions evolve over time.

Watermarking provides a provenance signal embedded at generation time, enabling later detection or attribution.
Recent work optimises watermark trade-offs between detection strength, quality impact, and robustness \citep{Wouters24-Watermarks}.
At the same time, impossibility results formalise limits on ``strong'' watermarking that would remain unremovable without quality loss \citep{Zhang24-StrongWatermarking}.
These findings reinforce the need for rigorous evaluation of non-provenance detectors under realistic shift.

Taken together, the literature converges on a common message: strong in-domain performance does not guarantee reliable behaviour under distribution shift.
\section{Datasets}
\label{sec:datasets}

We evaluate supervised binary AI-text detection under distribution shift using three complementary benchmarks:
(i) \dataset{HC3~PLUS} for in-domain QA and semantic-invariant rewrites,
(ii) \dataset{M4} for cross-dataset transfer across domains and generators (English subset),
and (iii) \dataset{AI-Text-Detection-Pile} for large-scale external evaluation.

\subsection{\dataset{HC3~PLUS}}
\label{subsec:hc3plus}

\dataset{HC3} pairs human answers with ChatGPT outputs for the same prompts \citep{Guo23-HC3}.
\dataset{HC3~PLUS} extends this setting with semantic-invariant transformations (e.g., paraphrasing, summarisation, translation),
which better reflect post-edited or re-expressed AI text in deployment \citep{Su23-HC3Plus}.
All AI responses are generated by ChatGPT (\model{GPT-3.5-Turbo}) \citep{Su23-HC3Plus}.
Split sizes are reported in Appendix~\ref{app:dataset_stats}.

\subsection{\dataset{Multi-Domain Multi-Generator M4}}
\label{subsec:m4}

\dataset{M4} is a black-box benchmark designed to stress-test generalisation across domains and generator families \citep{Wang24-M4}.
We use the English subset (78{,}766 samples; 12{,}583 Human / 66{,}183 AI) and report results by domain
(\texttt{wikipedia}, \texttt{wikihow}, \texttt{reddit}, \texttt{arxiv}, \texttt{peerread})
and by generator (eight families including \texttt{ChatGPT}, \texttt{Davinci}, \texttt{Cohere}, \texttt{BLOOMZ}, \texttt{Dolly}, \texttt{Dolly2}, \texttt{FlanT5}, \texttt{LLaMA}).
The domain and generator breakdowns are provided in Appendix~\ref{app:dataset_stats}.

\subsection{\dataset{AI-Text-Detection-Pile}}
\label{subsec:pile}

For external evaluation beyond \dataset{HC3~PLUS} and \dataset{M4}, we test on \dataset{AI-Text-Detection-Pile},
a large-scale out-of-distribution dataset spanning diverse human sources and AI-generated text \citep{Artem9k24-AITextDetectionPile}.
Dataset statistics are given in Appendix~\ref{app:dataset_stats}.
\section{Method}
\label{sec:method}
\begin{figure*}[t]
  \centering
  \includegraphics[width=\textwidth]{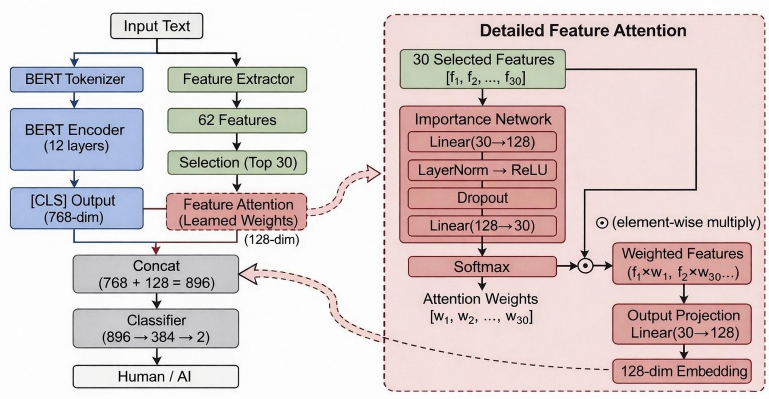}
  \caption{\textbf{\model{BERT+FeatAttn} architecture.}
  The input text is encoded by the transformer, yielding a \texttt{[CLS]} representation.
  In parallel, linguistic features are extracted, reduced to the selected subset, and mapped to a compact embedding via a feature-attention module.
  The \texttt{[CLS]} vector and feature embedding are concatenated and passed to an MLP classifier to predict Human vs AI.}
  \label{fig:bert_featattn}
\end{figure*}

We study supervised binary detection of AI-generated text under realistic distribution shift, i.e., when the test data differ from the training data in domain, writing style, or generator family.
Given an input text $x$, the detector outputs class probabilities
$p_\theta(y{=}0 \mid x)$ for human-written text and $p_\theta(y{=}1 \mid x)$ for AI-generated text, where $y\in\{0,1\}$.
In what follows we write $p_\theta(x)\equiv p_\theta(y{=}1 \mid x)\in[0,1]$ for brevity.
Our approach combines (i) strong transformer baselines and (ii) feature-augmented variants that fuse handcrafted linguistic signals with transformer representations via a lightweight feature-attention module.
To reflect deployment settings where target-domain labels are unavailable, we calibrate a single global decision threshold on \HCThreePlus{} validation and keep it fixed for all evaluations.

\subsection{Problem Setup and Metric}
\label{subsec:problem_setup}

For a fixed decision threshold $\tau$, predictions are obtained by
\begin{equation}
\yhat(x;\tau)=\mathbb{I}\big[p_\theta(x)\ge \tau\big].
\end{equation}

Since target distributions may be class-imbalanced and errors under shift can be asymmetric, we use
\emph{balanced accuracy} (BA) as the primary metric:
\begin{equation}
\ba=\tfrac{1}{2}\left(\tpr+\tnr\right),
\end{equation}
where $\tpr$ is the recall on the AI class (true positive rate) and $\tnr$ is the recall on the Human class (true negative rate).

\subsection{Base Transformer Detectors}
\label{subsec:base_detectors}

As baselines, we fine-tune pretrained transformer encoders for sequence classification (e.g., \model{BERT}, \model{RoBERTa})
\citep{Devlin19-BERT,Liu19-RoBERTa}.
Let $h_{\texttt{[CLS]}}(x)\in\mathbb{R}^{d}$ denote the pooled representation of the input.
A linear head predicts the AI probability:
\begin{equation}
p_\theta(x)=\sigma\!\big(W\,h_{\texttt{[CLS]}}(x)+b\big).
\end{equation}
All models are trained end-to-end with standard cross-entropy loss.

\subsection{Handcrafted Features and Global Selection}
\label{subsec:handcrafted_features}

To complement neural representations with explicit stylistic cues, we extract a handcrafted feature vector
$f(x)\in\mathbb{R}^{m}$ per text ($m{=}62$), spanning lexical diversity, POS/stylometric statistics, readability and punctuation measures, and LM-based signals such as perplexity and burstiness.
For the feature-augmented detectors, we retain a compact subset of $k{=}30$ features selected once on the source dataset to avoid any target-domain leakage.
Concretely, we compute a global importance ranking on the \HCThreePlus{} training set by combining mutual information and absolute point-biserial correlation with the binary labels, and select the top-$k$ features under this ranking.
The selected subset is then fixed for all samples and all test distributions.
Full definitions, normalisation, and the exact scoring procedure are reported in Appendix~\ref{app:features}.

\subsection{Feature-Attention Fusion Detectors}
\label{subsec:feat_fusion_models}

We define feature-augmented detectors that fuse transformer representations with a learnt, attention-weighted feature embedding.
The design is dual-branch:
(i) a text branch producing $h_{\texttt{[CLS]}}(x)$ and
(ii) a feature branch that learns \emph{dynamic} per-sample weights over $f_k(x)$ and projects the weighted features into a compact embedding.

\paragraph{Feature attention (dynamic gating).}
We compute attention weights over the selected features via a small importance network:
\begin{align}
u(x) &= W_2\,\phi\!\left(\mathrm{LN}\!\left(W_1 f_k(x)\right)\right), \\
a(x) &= \mathrm{softmax}\!\big(u(x)\big),
\end{align}
where $W_1\in\mathbb{R}^{128\times k}$, $W_2\in\mathbb{R}^{k\times 128}$, $\mathrm{LN}$ is layer normalisation,
and $\phi$ is ReLU (with dropout during training).
We form a weighted feature vector $\bar f_k(x)=a(x)\odot f_k(x)$ and project it to a feature embedding:
\begin{equation}
z_f(x)=W_3\,\bar f_k(x)\in\mathbb{R}^{128},
\qquad
W_3\in\mathbb{R}^{128\times k}.
\end{equation}
Unlike the Top-$k$ selection (static), this attention is \textbf{dynamic} and can vary across samples.

\paragraph{Fusion and prediction.}
We concatenate the transformer representation and the feature embedding and apply an MLP classifier:
\begin{equation}
h(x)=[\,h_{\texttt{[CLS]}}(x);\; z_f(x)\,],
\qquad
p_\theta(x)=\sigma\!\big(\mathrm{MLP}(h(x))\big).
\end{equation}

\begin{figure*}[t]
  \centering
  \includegraphics[width=\textwidth]{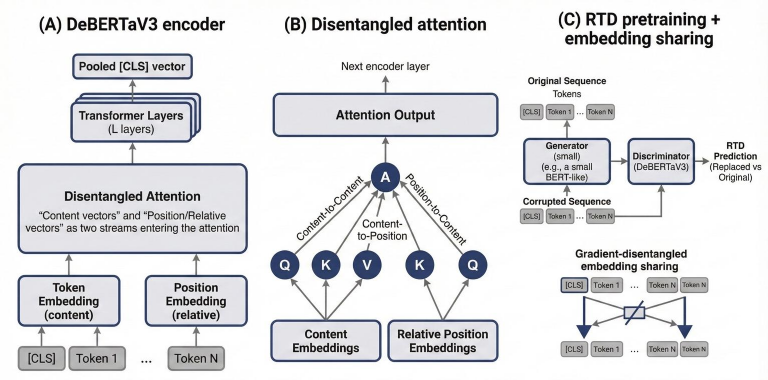}
  \caption{\textbf{DeBERTa-v3 overview.}
  (A) Encoder block using disentangled attention with separate content and relative-position representations.
  (B) Disentangled attention factorises content-to-content, content-to-position, and position-to-content interactions.
  (C) ELECTRA-style pre-training via replaced-token detection (RTD) with gradient-disentangled embedding sharing.}
  \label{fig:deberta_featattn}
\end{figure*}

\paragraph{Instantiations.}
We instantiate the fusion detector with two backbones:
\model{BERT+FeatAttn} (built on \model{BERT} \citep{Devlin19-BERT}) and
\model{DeBERTa+FeatAttn} (built on \model{DeBERTa} \citep{He20-DeBERTa}, using the \texttt{deberta-v3-base} family \citep{He21-DeBERTaV3}).
They share the same handcrafted feature pipeline and feature-attention module, differing only in the text encoder.
Figures~\ref{fig:bert_featattn} and~\ref{fig:deberta_featattn} illustrate the two architectures.

\paragraph{DeBERTa-v3 backbone.}
In addition to the DeBERTa-based fusion model, we also evaluate \model{DeBERTa-v3-base} as a standalone backbone.
DeBERTa-v3 uses ELECTRA-style replaced-token detection (RTD) during pre-training, which explicitly trains the encoder to distinguish replaced tokens from original ones \citep{He21-DeBERTaV3}.
We hypothesise that this objective may be beneficial when superficial cues vary due to rewriting or cross-domain shift.

\subsection{Threshold Calibration (Single Global Threshold)}
\label{subsec:threshold_calib}

To reflect deployment settings where target-domain labels are unavailable, we calibrate a \emph{single global} decision
threshold \emph{per detector} on held-out \HCThreePlus{} validation data and keep it fixed for all downstream evaluations.
For each detector, we pool the two validation views,
$\mathcal{D}_{\mathrm{val}} = \texttt{val\_qa} \cup \texttt{val\_si}$, and select a threshold from a fixed grid
$\mathcal{T}\subset[0,1]$ by maximising balanced accuracy on $\mathcal{D}_{\mathrm{val}}$:
\begin{equation}
\tau^{\star}
=
\arg\max_{\tau\in\mathcal{T}}
\mathrm{BA}_{\mathcal{D}_{\mathrm{val}}}(\tau).
\end{equation}
Once selected, $\tau^{\star}$ is kept fixed for all test distributions (no target-domain fine-tuning and no
threshold re-calibration), enabling a deployment-motivated assessment of robustness under distribution shift.
Grid specification (range and step) is reported in Appendix~\ref{app:threshold}.
\section{Experimental Protocol}
\label{sec:experiments}

We evaluate supervised binary AI-generated text detection under realistic \emph{distribution shift}.
All detectors are fine-tuned on \dataset{HC3~PLUS} \citep{Su23-HC3Plus} and assessed in three settings:
(i) in-domain testing on \dataset{HC3~PLUS},
(ii) held-out cross-dataset transfer to \dataset{M4} (English) \citep{Wang24-M4} with domain- and generator-level reporting,
and (iii) external evaluation on \dataset{AI-Text-Detection-Pile} \citep{Artem9k24-AITextDetectionPile}.
To reflect deployment conditions where target-domain labels are unavailable for calibration, we use a
validation-calibrated \emph{fixed-threshold} protocol throughout.

\subsection{Models Evaluated}
\label{subsec:models_evaluated}
We compare two standard transformer baselines, \model{BERT} \citep{Devlin19-BERT} and \model{RoBERTa} \citep{Liu19-RoBERTa},
against two feature-augmented detectors that fuse handcrafted linguistic features with transformer representations via
a feature-attention module: \model{BERT+FeatAttn} and \model{DeBERTa+FeatAttn} \citep{He20-DeBERTa}.
Architectural details are given in Section~\ref{sec:method}.

\subsection{Training and Validation-Calibrated Thresholding}
\label{subsec:train_calib}
Each detector is fine-tuned on the official \dataset{HC3~PLUS} \texttt{train} split.
We then calibrate a single decision threshold \emph{per detector} on the pooled validation set
$\mathcal{D}_{\mathrm{val}}=\texttt{val\_qa}\cup\texttt{val\_si}$.
Given predicted probabilities $p_\theta(x)\in[0,1]$ for the AI class, we convert scores to binary decisions using a fixed
threshold $\tau$ (predict AI if $p_\theta(x)\ge\tau$).
We select $\tau^\star$ by grid search over a uniform set of candidate thresholds $\mathcal{T}\subset[0,1]$ to maximise balanced
accuracy on $\mathcal{D}_{\mathrm{val}}$ (grid range and step are specified in Appendix~\ref{app:threshold}).
Once selected, $\tau^\star$ is \emph{kept fixed} for all subsequent test distributions, with no target-domain adaptation.

\subsection{Evaluation Suites}
\label{subsec:evaluation_suites}

We assess model robustness across three complementary evaluation suites that together cover the main challenges encountered in real-world deployment.

\paragraph{In-domain (\dataset{HC3~PLUS}).}
We report results on \texttt{test\_qa} (question answering) and \texttt{test\_si} (semantic-invariant rewrites), using the
same calibrated threshold $\tau^\star$.

\paragraph{Cross-dataset (\dataset{M4}, English; held-out target).}
We evaluate \dataset{HC3~PLUS}-trained detectors on the five English domains of \dataset{M4} and report:
(i) \emph{domain-level} performance (macro-averaged across domains), and
(ii) \emph{generator-level} performance by aggregating predictions by generator label.
No target-domain fine-tuning or threshold re-calibration is performed.

\paragraph{External (\dataset{AI-Text-Detection-Pile}).}
Finally, we evaluate on \dataset{AI-Text-Detection-Pile} as a large-scale external benchmark to assess robustness beyond
the two primary datasets, again using the same fixed threshold $\tau^\star$.

\subsection{Metrics and Reporting}
\label{subsec:metrics_reporting_experiments}
We use balanced accuracy (BA; defined in Section~\ref{subsec:problem_setup}) as the primary metric, as it remains
informative under class imbalance and asymmetric failure modes.
Where informative (notably for external evaluation and generator-level analysis), we additionally report accuracy, macro-F1,
ROC-AUC, and class-wise recall (Human/AI) to make systematic biases explicit.

\paragraph{Implementation details.}
Optimiser settings, learning-rate schedule, batch size, maximum sequence length, and training epochs are reported in
Appendix~\ref{app:training}.
\section{Results}
\label{sec:results}

We organise results to separate three sources of difficulty: semantic-invariant rewriting within \dataset{HC3~PLUS}, dataset/domain shift on \dataset{M4}, and external shift on \dataset{AI-Text-Detection-Pile}.
Throughout, each model uses a fixed threshold calibrated on \dataset{HC3~PLUS} validation (Section~\ref{sec:experiments}), and we highlight how robustness depends on both the target domain and generator family.

\subsection{In-domain Performance on \dataset{HC3~PLUS} (Baselines)}
\label{subsec:hc3_indomain}

We begin with two transformer baselines (\model{BERT} and \model{RoBERTa}) fine-tuned on the official
\dataset{HC3~PLUS} \texttt{train} split.
Following the fixed-threshold protocol (Section~\ref{sec:experiments}), we calibrate one threshold $\tau^\star$ per detector
on the pooled validation set $\texttt{val\_qa}\cup\texttt{val\_si}$, and keep it fixed when evaluating on
\texttt{test\_qa} and \texttt{test\_si}.
Table~\ref{tab:hc3_indomain_ba} reports balanced accuracy (BA) as our primary metric.

\begin{center}
\footnotesize
\setlength{\tabcolsep}{4pt}
\begin{tabular}{lccr}
\toprule
Model & $\tau^\star$ & BA \texttt{test\_qa} & BA \texttt{test\_si} \\
\midrule
\model{BERT}    & 0.72 & 98.26 & 85.97 \\
\model{RoBERTa} & 0.76 & \textbf{99.54} & \textbf{86.58} \\
\bottomrule
\end{tabular}
\captionof{table}{HC3~PLUS in-domain BA (\%) with fixed $\tau^\star$ (per detector).}
\label{tab:hc3_indomain_ba}
\end{center}

As shown in Table~\ref{tab:hc3_indomain_ba}, performance is near-ceiling on \texttt{test\_qa}, whereas the semantic-invariant split
\texttt{test\_si} is markedly more challenging, consistent with meaning-preserving transformations reducing reliance on surface-level cues.

\subsection{Cross-dataset Transfer to \dataset{M4}: Domain-level Results (Baselines)}
\label{subsec:m4_domains}

We next evaluate the same \dataset{HC3~PLUS}-trained baselines under dataset shift on the five English \dataset{M4} domains.
To characterise domain-dependent behaviour under a fixed operating point, Table~\ref{tab:m4_domain_transfer} reports BA together
with class-wise recall (H-R, AI-R).

\begin{center}
\small
\setlength{\tabcolsep}{4pt}
\begin{tabular}{l ccc ccc}
\toprule
& \multicolumn{3}{c}{\model{BERT}} & \multicolumn{3}{c}{\model{RoBERTa}} \\
Domain & BA & H-R & AI-R & BA & H-R & AI-R \\
\midrule
\texttt{wikipedia} & \textbf{87.4} & 90.1 & 84.7 & 68.9 & 48.4 & 89.3 \\
\texttt{wikihow}   & \textbf{80.7} & 93.4 & 68.1 & 65.0 & 50.8 & 79.3 \\
\texttt{reddit}    & \textbf{85.8} & 99.1 & 72.4 & 73.8 & 53.9 & 93.8 \\
\texttt{arxiv}     & 69.4 & 58.5 & 80.2 & \textbf{92.3} & 100.0 & 84.6 \\
\texttt{peerread}  & 74.9 & 100.0 & 49.8 & \textbf{89.3} & 99.3 & 79.3 \\
\midrule
\textbf{Macro Avg.} & \textbf{79.6} & 88.2 & 71.0 & 77.9 & 70.5 & 85.3 \\
\bottomrule
\end{tabular}
\captionof{table}{M4 domain transfer (English): BA, H-R, AI-R (\%). Best BA per row is in bold.}
\label{tab:m4_domain_transfer}
\end{center}

The transfer results in Table~\ref{tab:m4_domain_transfer} are strongly domain-dependent and reveal complementary error profiles under shift.
Across several domains, \model{BERT} achieves higher H-R (i.e., it is more likely to label samples as Human when appropriate),
whereas \model{RoBERTa} often achieves higher AI-R (i.e., it more readily labels samples as AI), sometimes at the cost of lower H-R.

\subsection{External Evaluation on \dataset{AI-Text-Detection-Pile}}
\label{subsec:pile_eval}

Finally, we evaluate \model{BERT} and \model{RoBERTa} (trained on \dataset{HC3~PLUS}) on \dataset{AI-Text-Detection-Pile}.
We keep thresholds fixed from \dataset{HC3~PLUS} validation and report BA together with class-wise recall and additional metrics
(Table~\ref{tab:pile_models}).

\begin{center}
\small
\setlength{\tabcolsep}{4pt}
\begin{tabular}{l c c c c c c}
\toprule
Model & BA & Acc & H-R & AI-R & ROC-AUC & F1 \\
\midrule
\model{BERT}    & \textbf{75.73} & \textbf{82.19} & \textbf{89.26} & 62.20 & \textbf{87.60} & \textbf{64.61} \\
\model{RoBERTa} & 64.67 & 52.67 & 39.52 & \textbf{89.81} & 80.68 & 49.80 \\
\bottomrule
\end{tabular}
\captionof{table}{Pile external: BA / Acc / H-R / AI-R / ROC-AUC / F1 (\%).}
\label{tab:pile_models}
\end{center}

Table~\ref{tab:pile_models} highlights complementary biases under external shift: \model{BERT} attains substantially higher H-R,
indicating a stronger tendency to classify Human text correctly, whereas \model{RoBERTa} attains higher AI-R, indicating a stronger
tendency to classify samples as AI.

\begin{figure}[t]
  \centering
  \includegraphics[width=\linewidth]{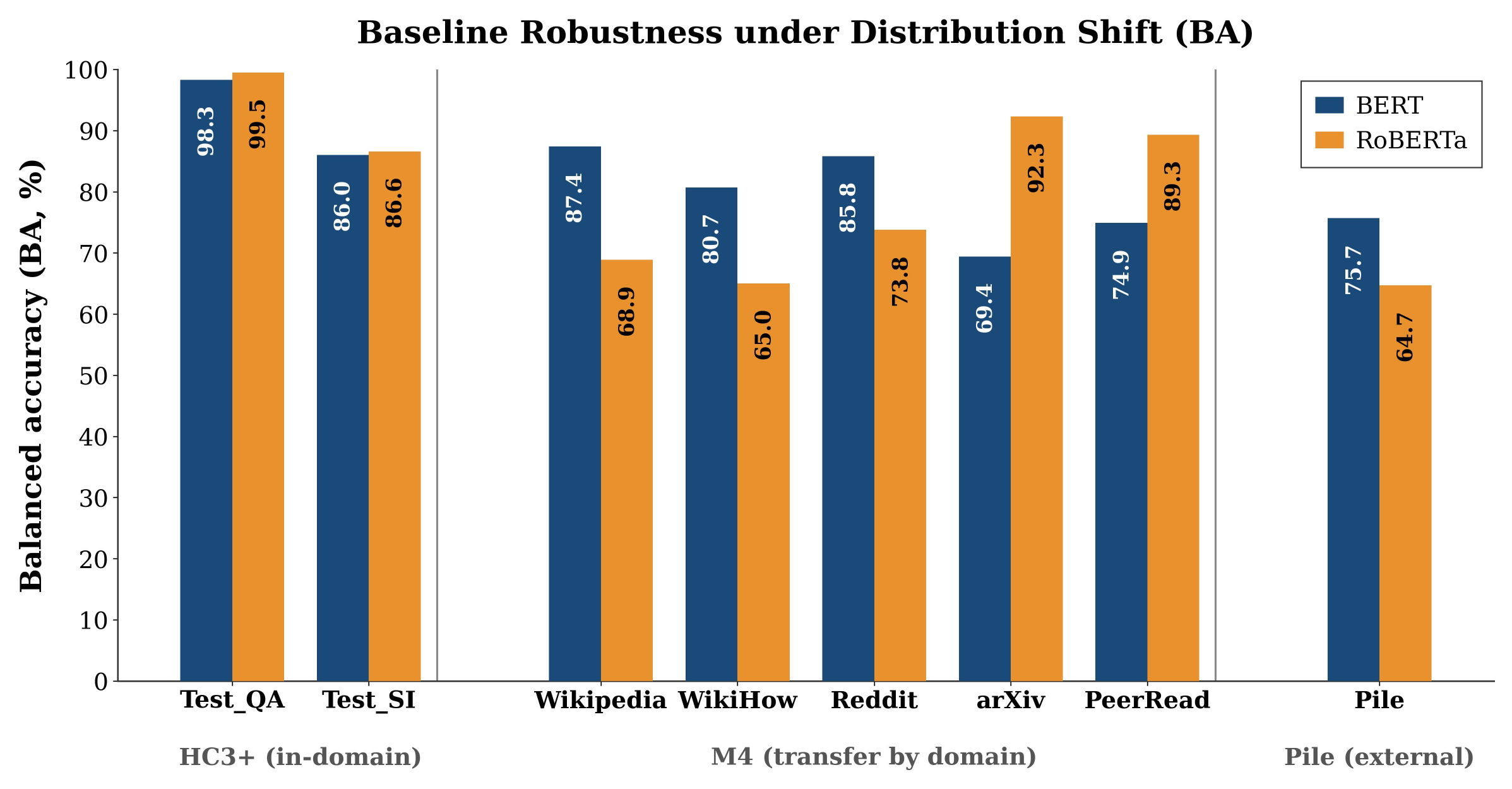}
  \caption{\textbf{Baseline robustness under distribution shift (BA).}
  Balanced accuracy (BA) of \model{BERT} and \model{RoBERTa} under a fixed decision threshold calibrated on \dataset{HC3~PLUS} validation and held constant across targets.
  Results are shown for \dataset{HC3~PLUS} (\texttt{test\_qa}, \texttt{test\_si}), \dataset{M4} English domains, and \dataset{AI-Text-Detection-Pile}.}
  \label{fig:baseline_robustness_ba}
\end{figure}

\subsection{In-domain Performance on \dataset{HC3~PLUS} (Feature-augmented Detectors)}
\label{subsec:hc3_featattn}

We now turn to feature-augmented detectors that fuse 30 selected linguistic features with transformer representations
via a feature-attention module (Section~\ref{sec:method}): \model{BERT+FeatAttn} and \model{DeBERTa+FeatAttn}.
Thresholds are again calibrated on $\texttt{val\_qa}\cup\texttt{val\_si}$ and fixed for evaluation.
Table~\ref{tab:adv_hc3} reports in-domain BA.

\begin{center}
\footnotesize
\setlength{\tabcolsep}{4pt}
\begin{tabular}{lccr}
\toprule
Model & $\tau^\star$ & BA \texttt{test\_qa} & BA \texttt{test\_si} \\
\midrule
\model{BERT+FeatAttn}     & 0.70 & 98.83 & 86.16 \\
\model{DeBERTa+FeatAttn}  & 0.77 & \textbf{99.47} & \textbf{87.21} \\
\bottomrule
\end{tabular}
\captionof{table}{HC3~PLUS in-domain BA (\%) for feature-augmented detectors.}
\label{tab:adv_hc3}
\end{center}

As shown in Table~\ref{tab:adv_hc3}, \model{DeBERTa+FeatAttn} improves BA on both splits at the calibrated operating point.

\subsection{Cross-dataset Transfer to \dataset{M4}: Domain-level Results (Feature-augmented Detectors)}
\label{subsec:m4_domains_adv}

We next evaluate the same feature-augmented detectors under direct transfer to \dataset{M4}.
Table~\ref{tab:adv_m4_domains} reports BA by domain.

\begin{center}
\small
\setlength{\tabcolsep}{6pt}
\begin{tabular}{lcc}
\toprule
Domain & \model{BERT+FeatAttn} & \model{DeBERTa+FeatAttn} \\
\midrule
\texttt{wikipedia} & \textbf{89.96} & 74.52 \\
\texttt{wikihow}   & \textbf{81.31} & 79.47 \\
\texttt{reddit}    & 86.36 & \textbf{90.23} \\
\texttt{arxiv}     & \textbf{75.55} & 74.69 \\
\texttt{peerread}  & 76.33 & \textbf{93.00} \\
\midrule
\textbf{Macro Avg.} & 81.90 & \textbf{82.39} \\
\bottomrule
\end{tabular}
\captionof{table}{M4 domain transfer: feature-augmented BA (\%).}
\label{tab:adv_m4_domains}
\end{center}

Table~\ref{tab:adv_m4_domains} shows pronounced domain dependence under shift, while feature augmentation improves macro-average transfer
and changes which backbone performs best depending on the target domain.

\begin{figure}[H]
  \centering
  \includegraphics[width=\linewidth]{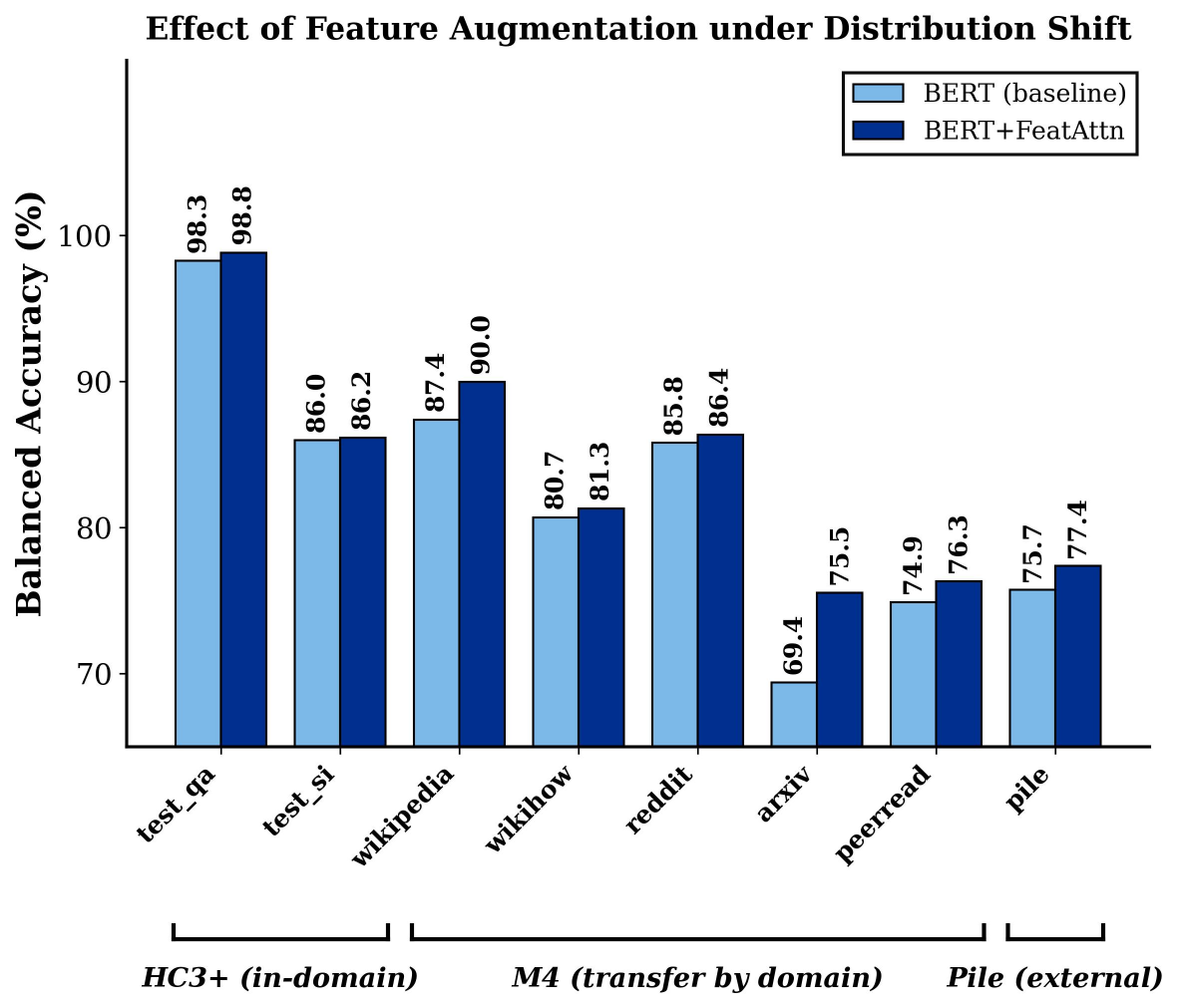}
  \caption{\textbf{Effect of feature augmentation under distribution shift (BA).}
  Balanced accuracy (BA) of \model{BERT} versus \model{BERT+FeatAttn} under the same fixed threshold calibrated on \dataset{HC3~PLUS} validation.
  We report in-domain performance on \dataset{HC3~PLUS} (\texttt{test\_qa}, \texttt{test\_si}), domain-level transfer on \dataset{M4}, and external evaluation on \dataset{AI-Text-Detection-Pile}.
  Feature augmentation improves robustness across all targets, with the largest gain on \texttt{arXiv} (+6.15 BA points).}
  \label{fig:bert_featattn_shift_ba}
\end{figure}

\subsection{\dataset{M4} Error Profiling by Generator}
\label{subsec:m4_generators}

Beyond domain shift, \dataset{M4} also varies generator families and generation pipelines.
To isolate generator-specific behaviour, Table~\ref{tab:adv_m4_generators} reports class-wise recall computed per generator:
the Human row corresponds to recall on human-written text (H-R), and each generator row corresponds to recall on that AI source (AI-R).

\begin{center}
\small
\setlength{\tabcolsep}{6pt}
\begin{tabular}{lccr}
\toprule
Generator & \model{BERT+FeatAttn} & \model{DeBERTa+FeatAttn} \\
\midrule
\texttt{Human}   & \textbf{92.95} & 69.42 \\
\texttt{ChatGPT} & 98.23 & \textbf{99.86} \\
\texttt{Cohere}  & 90.20 & \textbf{98.20} \\
\texttt{Davinci} & 84.74 & \textbf{95.53} \\
\texttt{Dolly}   & 65.43 & \textbf{86.98} \\
\texttt{Dolly2}  & 48.83 & \textbf{69.83} \\
\texttt{Bloomz}  & 29.83 & \textbf{79.84} \\
\texttt{Llama}   & 12.80 & \textbf{95.22} \\
\texttt{FlanT5}  & 80.27 & \textbf{93.22} \\
\bottomrule
\end{tabular}
\captionof{table}{M4 generator detection rates: feature-augmented (class-wise recall, \%).}
\label{tab:adv_m4_generators}
\end{center}

Table~\ref{tab:adv_m4_generators} reveals a clear trade-off: \model{DeBERTa+FeatAttn} generalises substantially better across diverse
(open and proprietary) generators, whereas \model{BERT+FeatAttn} preserves human text more reliably.
This motivates class-wise reporting and generator-aware breakdowns rather than relying on a single aggregate score.

\subsection{DeBERTa-v3 Optimisation for Balanced Transfer}
\label{subsec:deberta_opt}

To further improve robustness under distribution shift and obtain a more balanced error profile, we conducted an optimisation study around \model{DeBERTa-v3-base} with feature-attention.
The best configuration uses 30 selected linguistic features, a weighted loss (Human: 1.5; AI: 1.0), and a fixed validation-calibrated threshold $\tau^\star=0.60$.
Table~\ref{tab:deberta_opt_overall} summarises in-domain and cross-dataset performance, while Table~\ref{tab:deberta_opt_generators} reports generator-wise recall on \dataset{M4}.

\begin{center}
\footnotesize
\setlength{\tabcolsep}{4pt}
\renewcommand{\arraystretch}{1.05}
\begin{tabular}{lcccc}
\toprule
Test set & BA & H-R & AI-R & ROC-AUC \\
\midrule
HC3 QA   & \textbf{99.63}\,\% & 99.37\,\% & 99.90\,\% & 1.0000 \\
HC3 SI   & \textbf{86.85}\,\% & 81.81\,\% & 91.88\,\% & 0.9566 \\
M4 (All) & \textbf{85.97}\,\% & 81.38\,\% & 90.57\,\% & 0.9437 \\
\bottomrule
\end{tabular}
\captionof{table}{\model{DeBERTa-v3-base+FeatAttn} best setting (30 features; Human weight 1.5; $\tau^\star{=}0.60$): overall performance (all values in \%).}
\label{tab:deberta_opt_overall}
\end{center}

\begin{center}
\footnotesize
\setlength{\tabcolsep}{6pt}
\renewcommand{\arraystretch}{1.05}
\begin{tabular}{lrr}
\toprule
Generator & Samples & Recall \\
\midrule
\texttt{Human}   & 12{,}583 & 81.38\,\% \\
\texttt{ChatGPT} & 12{,}581 & 99.80\,\% \\
\texttt{LLaMA}   & 586      & 96.25\,\% \\
\texttt{Bloomz}  & 12{,}000 & 94.59\,\% \\
\texttt{FlanT5}  & 6{,}000  & 93.97\,\% \\
\texttt{Cohere}  & 10{,}142 & 93.46\,\% \\
\texttt{Davinci} & 12{,}586 & 89.35\,\% \\
\texttt{Dolly2}  & 3{,}000  & 76.20\,\% \\
\texttt{Dolly}   & 9{,}288  & 73.45\,\% \\
\bottomrule
\end{tabular}
\captionof{table}{Generator-wise recall on \dataset{M4} for the best \model{DeBERTa-v3-base+FeatAttn} setting. The \texttt{Human} row is Human recall; all other rows are AI recall (in \%).}
\label{tab:deberta_opt_generators}
\end{center}

The resulting profile is both strong and comparatively balanced under shift: in-domain performance is near-ceiling on HC3 QA (99.63\% BA) and remains robust on semantic-invariant rewrites (86.85\% BA). On M4, the model reaches 85.97\% BA with 81.38\% Human recall and 90.57\% AI recall, while achieving notably high detection on the hardest generator Bloomz (94.59\%).

\subsection{Multi-Seed Stability and Stronger Baselines}
To ensure the reliability and reproducibility of our results, we re-evaluated our best model (\model{DeBERTa-v3-base+FeatAttn}) across 5 independent random seeds. Results remain highly stable across all datasets and metrics:

\FloatBarrier
\begin{table}[!htbp]
\centering
\small
\begin{tabular}{lccc}
\toprule
Dataset & BA (\%) & H-R (\%) & AI-R (\%) \\
\midrule
HC3 QA     & 99.67 $\pm$ 0.04 & 99.45 $\pm$ 0.06 & 99.90 $\pm$ 0.03 \\
HC3 SI     & 87.34 $\pm$ 0.15 & 84.71 $\pm$ 1.61 & 90.00 $\pm$ 1.51 \\
M4 (macro) & 83.15 $\pm$ 1.04 & 80.05 $\pm$ 4.63 & 86.24 $\pm$ 2.85 \\
\bottomrule
\end{tabular}
\caption{Multi-seed results for \model{DeBERTa-v3-base+FeatAttn} (mean $\pm$ std across 5 seeds). BA = Balanced Accuracy, H-R = Human Recall, AI-R = AI Recall.}
\label{tab:multiseed}
\end{table}
\FloatBarrier

We also evaluated four strong external baselines under the exact same fixed-threshold protocol to provide a broader comparison:

\FloatBarrier
\begin{table}[!htbp]
\centering
\begin{tabular}{lc}
\toprule
Method & BA (\%) \\
\midrule
DeBERTa-v3+FeatAttn (ours) & \textbf{84.50} \\
Fast-DetectGPT             & 77.28 \\
Log-Rank (GPT-2)           & 73.17 \\
RADAR (NeurIPS 2023)       & 67.82 \\
\bottomrule
\end{tabular}
\caption{M4 macro-average balanced accuracy under the same fixed-threshold protocol.}
\label{tab:baselines}
\end{table}
\FloatBarrier

Our model outperforms the strongest zero-shot baseline by +7.22 percentage points.

\subsection{Feature-Category Ablations}
To understand which linguistic signals drive robustness under distribution shift, we performed a complete category-level ablation on the 30 selected features while keeping the operating point fixed. This analysis isolates the contribution of each feature group rather than individual features.

\FloatBarrier
\begin{table}[!htbp]
\centering
\begin{tabular}{lcc}
\toprule
Category (single only) & M4 BA (\%) & Human Recall (\%) \\
\midrule
Readability   & \textbf{83.15} & 89.80 \\
Coherence     & 82.94 & 80.16 \\
Repetition    & 81.73 & 74.85 \\
Perplexity    & 80.81 & 71.89 \\
\bottomrule
\end{tabular}
\caption{Single-category-only performance on M4 (macro average).}
\label{tab:single_category}
\end{table}
\FloatBarrier

\FloatBarrier
\begin{table}[!htbp]
\centering
\begin{tabular}{lcc}
\toprule
Removed Category & $\Delta$ M4 BA & Interpretation \\
\midrule
Readability  & -1.65 & Largest overall drop \\
Vocabulary   & -1.52 & Strongest H-R impact \\
Stylometric  & -1.04 & Moderate effect \\
Perplexity   & -0.79 & Shifts operating point \\
\bottomrule
\end{tabular}
\caption{Leave-one-category-out results on M4 (change from full 30-feature model).}
\label{tab:drop_category}
\end{table}
\FloatBarrier

These ablations confirm that \textbf{Readability} and \textbf{Vocabulary} features are the most critical for achieving robust cross-domain performance.

\subsection{External Evaluation on \dataset{AI-Text-Detection-Pile} and Gemini Generator Shift}
\label{subsec:pile_eval_adv}

We next evaluate our best feature-augmented configuration under external shift, keeping the validation-calibrated operating point fixed ($\tau^*=0.60$). On \dataset{AI-TEXT-DETECTION-PILE}, \model{DeBERTa-v3-base+FeatAttn} (30 features; Human loss weight 1.5) attains 71.86\% balanced accuracy, with 88.19\% AI recall and 55.53\% human recall (ROC-AUC 81.47\%, F1 56.20\%). This profile indicates strong sensitivity to AI text under shift, but also a substantial false-positive rate on human text at the same fixed threshold.

To probe generator shift beyond the benchmark generators, we additionally evaluate on Gemini outputs produced from the \dataset{HC3~PLUS} QA prompts. Specifically, we regenerate answers for the same questions using \textbf{Gemini 2.0 Flash}. On these 23,463 Gemini 2.0 Flash answers (all AI-generated), AI recall reaches \textbf{99.63\%}. We further consider a harder setting with 1,000 answers generated by \textbf{Gemini 3.0 Pro}, where AI recall drops to \textbf{82.70\%}. In summary, these results show that even with an unchanged detector and fixed operating point, transfer can vary substantially across generator families and generation setups.

We further test generalization on a large-scale academic-domain dataset called \textbf{Academic Text} \citep{mady2026academictext}. It consists of 469,008 human-written arXiv paragraphs (collected from papers published before 2022, guaranteeing purely human-generated text) and 200,000 AI-generated academic texts produced via the official APIs: 100,000 samples by GPT-3.5 Turbo and 100,000 samples by Gemini 2.0 Flash. The AI texts were generated on STEM topics carefully matched to the distribution of the human arXiv abstracts, ensuring close semantic and topical alignment. Under the same fixed threshold calibrated on HC3 PLUS validation ($\tau^*=0.60$), our model achieves an overall balanced accuracy of \textbf{81.09\%}, with perfect detection on GPT-3.5 Turbo (\textbf{100.0\%}) and strong performance on Gemini 2.0 Flash (\textbf{93.25\%}). Human recall on the arXiv text is 75.57\%.

In summary, these results show that even with an unchanged detector and fixed operating point, transfer can vary substantially across generator families, domains, and generation setups. The academic-domain evaluation provides additional evidence of robust real-world generalization under distribution shift.
\section{Discussion}
\label{sec:discussion}

Our experiments indicate that performance on the source distribution can be a weak proxy for reliability under shift in domain, writing style, and generator family. Evaluating the same fixed-threshold detectors on \dataset{HC3~PLUS}, held-out \dataset{M4}, and \dataset{AI-Text-Detection-Pile} surfaces error patterns that are not apparent from source-only reporting.

Semantic-invariant evaluation is a particularly stringent stress test. While results are near-ceiling on QA-style data, they drop on meaning-preserving rewrites, suggesting that detectors can exploit surface regularities that do not survive paraphrasing, summarisation, or translation. This gap matters in practice, where AI text is often post-edited or re-expressed before detection.

Holding the decision threshold fixed after validation calibration helps interpret these gaps under realistic constraints. Without re-tuning to each target distribution, transfer on \dataset{M4} becomes clearly heterogeneous across domains and generators. Multi-seed experiments confirm that our best model (\model{DeBERTa-v3-base+FeatAttn}) is highly stable (83.15 $\pm$ 1.04\% macro BA on M4), and it consistently outperforms strong zero-shot baselines (Fast-DetectGPT, Log-Rank, RADAR) by up to +7.22 points under the same protocol. Generator-wise reporting further shows that aggregate scores can conceal large swings across generator families, supporting generator-aware profiling alongside domain averages.

Feature augmentation via attention-based fusion improves transfer robustness in several cases, but it also sharpens the operating-point trade-off between human-text preservation and broad generator coverage. Our category-level ablations reveal that \textbf{Readability} and \textbf{Vocabulary} features contribute most to cross-domain robustness, while other categories have smaller or even detrimental effects under shift.

External evaluation on both \dataset{AI-Text-Detection-Pile} and the new large-scale \textbf{Academic Text} dataset further confirms that the fixed-threshold protocol reveals practically relevant failure modes. Even with an unchanged detector, performance varies substantially across generator families and domains, underscoring the importance of stress-testing under realistic deployment conditions.

Finally, class-wise reporting remains essential beyond the training distribution. External benchmarks can differ in class priors and error asymmetry, so accuracy alone can be misleading; balanced accuracy together with class-wise recall makes systematic biases explicit. Collectively, our results motivate semantic-invariant stress tests, fixed validation-calibrated operating points, generator-aware error profiling, and explicit feature analysis when assessing practical detector robustness.

\paragraph{Broader Impact and Ethical Considerations.}
This work aims to improve the robustness of AI-generated text detection in real-world deployment scenarios, supporting content moderation, academic integrity, and the fight against misinformation. By exposing operating-point trade-offs under a fixed-threshold protocol, we highlight practical risks that are often masked in standard evaluations. While our approach strengthens detection reliability, we acknowledge potential dual-use concerns: adversaries could exploit the identified weaknesses to craft harder-to-detect text, and overly aggressive detectors risk false positives that could suppress legitimate human writing. We therefore advocate for transparent, deployment-realistic evaluation and responsible use of such detectors in high-stakes applications.
\section{Conclusion}
\label{sec:conclusion}

We studied supervised binary AI-text detection under distribution shift using a deployment-oriented fixed-threshold protocol.
Detectors were trained on \dataset{HC3~PLUS}, calibrated once on held-out validation, and then evaluated without target adaptation on semantic-invariant rewrites, multi-domain and multi-generator transfer in \dataset{M4}, and external testing on \dataset{AI-Text-Detection-Pile}.

Across models, near-ceiling in-domain results do not reliably predict performance under shift.
Semantic-invariant rewrites act as a strong stress test, and cross-dataset transfer exposes domain- and generator-specific failure modes that are obscured by aggregate reporting.
Feature augmentation improves robustness to generator variation, but it sharpens the trade-off between preserving human text and maximising AI recall.
External evaluation further highlights the importance of balanced and class-wise metrics for making asymmetric errors explicit.

Overall, these findings motivate evaluation protocols that stress semantic invariance, fix an operating point via held-out validation, and complement aggregate scores with domain- and generator-level analyses.

\clearpage

\ifdefined\icmlaccepted
\begin{ack}
We thank the anonymous reviewers for their constructive feedback.
We also thank colleagues and collaborators for helpful discussions.

\end{ack}
\fi

\bibliography{src/references}
\bibliographystyle{style/icml2026}

\clearpage
\appendix
\appendix

\section*{Appendix}
\label{app:appendix}

This appendix provides (i) dataset statistics and breakdowns used in our experiments, and (ii) additional details to facilitate reproducibility.

\section{Dataset Statistics}
\label{app:dataset_stats}

\subsection{\dataset{HC3~PLUS}: Split Sizes (English)}
\label{app:hc3plus_stats}

\begin{center}
\small
\setlength{\tabcolsep}{6pt}
\renewcommand{\arraystretch}{1.05}
\begin{tabular}{lrrr}
\toprule
Split & Human & AI & Total \\
\midrule
\texttt{train}     & 74{,}020 & 74{,}020 & 148{,}040 \\
\texttt{val\_qa}   & 2{,}919  & 2{,}920  & 5{,}839 \\
\texttt{val\_si}   & 2{,}919  & 2{,}920  & 5{,}839 \\
\texttt{test\_qa}  & 16{,}951 & 8{,}018  & 24{,}969 \\
\texttt{test\_si}  & 19{,}047 & 19{,}063 & 38{,}110 \\
\bottomrule
\end{tabular}
\captionof{table}{\dataset{HC3~PLUS} split sizes used in our experiments (English).}
\label{tab:hc3plus_hai}
\end{center}

\subsection{\dataset{M4} (English): Domain Breakdown}
\label{app:m4_domain_stats}

\begin{center}
\footnotesize
\setlength{\tabcolsep}{4pt}
\renewcommand{\arraystretch}{1.05}
\begin{tabularx}{\linewidth}{@{}l r r r@{}}
\toprule
Domain & Total & Human & AI \\
\midrule
\texttt{wikipedia} & 17{,}033 & 3{,}000 & 14{,}033 \\
\texttt{wikihow}   & 18{,}001 & 3{,}001 & 15{,}000 \\
\texttt{reddit}    & 19{,}220 & 3{,}000 & 16{,}220 \\
\texttt{arxiv}     & 20{,}996 & 2{,}996 & 18{,}000 \\
\texttt{peerread}  & 3{,}516  & 586     & 2{,}930 \\
\midrule
\textbf{Total}     & \textbf{78{,}766} & \textbf{12{,}583} & \textbf{66{,}183} \\
\bottomrule
\end{tabularx}
\captionof{table}{\dataset{M4} English domain sizes used in our evaluation \citep{Wang24-M4}.}
\label{tab:m4_domain_sizes}
\end{center}

\subsection{\dataset{M4} (English): Generator Breakdown}
\label{app:m4_generator_stats}

\begin{center}
\footnotesize
\setlength{\tabcolsep}{4pt}
\renewcommand{\arraystretch}{1.05}
\begin{tabularx}{\linewidth}{@{}l r l@{}}
\toprule
Generator (AI only) & Samples & Domain coverage \\
\midrule
\texttt{ChatGPT} & 12{,}581 & all five domains \\
\texttt{Davinci} & 12{,}586 & all five domains \\
\texttt{Cohere}  & 10{,}142 & all five domains \\
\texttt{BLOOMZ}  & 12{,}000 & four domains (not \texttt{peerread}) \\
\texttt{Dolly}   & 9{,}288  & four domains (not \texttt{wikihow}) \\
\texttt{Dolly2}  & 3{,}000  & \texttt{wikihow} only \\
\texttt{FlanT5}  & 6{,}000  & \texttt{reddit}, \texttt{arxiv} \\
\texttt{LLaMA}   & 586      & \texttt{peerread} only \\
\midrule
\textbf{Total AI} & \textbf{66{,}183} & \\
\bottomrule
\end{tabularx}
\captionof{table}{\dataset{M4} generator breakdown (English subset) for AI samples, with domain coverage notes \citep{Wang24-M4}.}
\label{tab:m4_generator_sizes}
\end{center}

\subsection{\dataset{AI-Text-Detection-Pile}: Dataset Size}
\label{app:pile_stats}

\noindent\dataset{AI-Text-Detection-Pile} contains 1{,}392{,}011 samples (Human: 1{,}028{,}144; AI: 363{,}867) \citep{Artem9k24-AITextDetectionPile}.

\subsection{Academic Text Dataset}
\label{app:academic_text}

We introduce the Academic Text dataset for evaluating robustness in the STEM/academic domain. It consists of 469,008 human-written arXiv paragraphs (collected from papers published before 2022, guaranteeing purely human-generated text), 100,000 AI-generated texts produced by GPT-3.5 Turbo, and 100,000 AI-generated texts produced by Gemini 2.0 Flash. The AI texts were generated on STEM topics carefully matched to the distribution of the human arXiv paragraphs, ensuring close semantic alignment. The total size is 669,008 samples. Under the fixed-threshold protocol ($\tau^*=0.60$), our best model achieves 81.09\% balanced accuracy overall, with 100.0\% detection on GPT-3.5 Turbo and 93.25\% on Gemini 2.0 Flash. Human recall on arXiv text is 75.57\%.

The dataset is publicly available at:
\url{https://huggingface.co/datasets/mohamedmady/Academic-Text-arxiv-gpt-gemini}

\section{Supplementary Experimental Analyses}
\subsection{Multi-Seed Stability}
We re-evaluated our best model across 5 random seeds. Results are highly stable:

\begin{table}[h]
\centering
\begin{tabular}{lccc}
\toprule
Dataset & BA (\%) & H-R (\%) & AI-R (\%) \\
\midrule
HC3 QA     & 99.67 $\pm$ 0.04 & 99.45 $\pm$ 0.06 & 99.90 $\pm$ 0.03 \\
HC3 SI     & 87.34 $\pm$ 0.15 & 84.71 $\pm$ 1.61 & 90.00 $\pm$ 1.51 \\
M4 (macro) & 83.15 $\pm$ 1.04 & 80.05 $\pm$ 4.63 & 86.24 $\pm$ 2.85 \\
\bottomrule
\end{tabular}
\caption{Multi-seed results for \model{DeBERTa-v3-base+FeatAttn}.}
\label{tab:multiseed_app}
\end{table}

\subsection{Comparison with Zero-Shot Baselines}
Under the identical fixed-threshold protocol on M4:

\begin{table}[h]
\centering
\begin{tabular}{lc}
\toprule
Method & BA (\%) \\
\midrule
DeBERTa-v3+FeatAttn & \textbf{84.50} \\
Fast-DetectGPT             & 77.28 \\
Log-Rank (GPT-2)           & 73.17 \\
RADAR (NeurIPS 2023)       & 67.82 \\
\bottomrule
\end{tabular}
\caption{M4 macro-average balanced accuracy.}
\label{tab:baselines_app}
\end{table}

\subsection{Feature-Category Ablations}
Full category-level results (single-only and leave-one-out) are shown in Tables 11 and 12 of the main paper. Readability and Vocabulary are consistently the most impactful categories.

\subsection{Fixed vs Adaptive Threshold}
On the challenging HC3 SI split, adaptive threshold re-calibration improves BA by only +0.48\%, but completely flips the error profile (higher Human recall, lower AI recall). This demonstrates that the fixed-threshold protocol reveals operationally relevant trade-offs that are hidden under optimistic per-target tuning.

\section{Additional Details for Reproducibility}
\label{app:reproducibility}

\subsection{Training Setup}
\label{app:training}

\subsubsection{Models and Hyperparameters}
All detectors are trained for \textbf{binary} classification (Human vs AI) with cross-entropy loss. For feature-augmented models, we jointly train the feature-attention module and an MLP classifier head together with the transformer backbone.

\subsubsection{Hardware and Software}
\label{app:hardware}
Experiments were run on an NVIDIA GeForce RTX 4090 GPU with \textbf{24\,GB VRAM}. Software stack: Python 3.10, PyTorch 2.x \citep{Paszke19-PyTorch}, HuggingFace Transformers 4.x \citep{Wolf20-Transformers}, CUDA 12.x, and scikit-learn 1.x \citep{Pedregosa11-ScikitLearn}.

\subsubsection{Training and Inference Time}
Typical runtime (3 epochs): BERT $\sim$1.5h, RoBERTa $\sim$1.5h, BERT+FeatAttn $\sim$2h, DeBERTa+FeatAttn $\sim$2.5h. Inference is $\sim$45--55 seconds per 1{,}000 samples depending on the backbone.

\subsection{Threshold Calibration}
\label{app:threshold}

\paragraph{Protocol.}
We select a single global threshold $\tau^\star$ on a pooled validation set (HC3~PLUS \texttt{val\_qa}$\cup$\texttt{val\_si}) via grid search, and keep it fixed for all test distributions to reflect deployment where target labels are unavailable.

\paragraph{Search details.}
We search $\tau \in [0.10, 0.90]$ with step size $0.01$ and optimise balanced accuracy:
\[
\mathrm{BA}=\tfrac{1}{2}(\mathrm{TPR}+\mathrm{TNR}).
\]
To prevent leakage, we use \emph{only} validation predictions to select $\tau^\star$.

\subsection{Handcrafted Linguistic Features}
\label{app:features}

\subsubsection{Feature Set Overview ($m{=}62$; $k{=}30$ selected)}
We extract a handcrafted feature vector $f(x)\in\mathbb{R}^{62}$ organised into eight categories, and select the top-$k$ features ($k{=}30$) for feature-attention models.

\begin{table*}[t]
\centering
\caption{Training hyperparameters for all models.}
\label{tab:hyperparams_all}
\small
\setlength{\tabcolsep}{5pt}
\renewcommand{\arraystretch}{1.05}
\resizebox{\textwidth}{!}{%
\begin{tabular}{@{}lcccc@{}}
\toprule
\textbf{Parameter} & \textbf{BERT} & \textbf{RoBERTa} & \textbf{BERT+FeatAttn} & \textbf{DeBERTa+FeatAttn} \\
\midrule
Base model & \texttt{bert-base-uncased} & \texttt{roberta-base} & \texttt{bert-base-uncased} & \texttt{microsoft/deberta-v3-base} \\
Hidden size $d$ & 768 & 768 & 768 & 768 \\
Layers / Heads & 12 / 12 & 12 / 12 & 12 / 12 & 12 / 12 \\
Max sequence length & 512 & 512 & 512 & 512 \\
Batch size & 16 & 16 & 16 & 16 \\
Epochs & 3 & 3 & 3 & 3 \\
Transformer LR & $2\times 10^{-5}$ & $2\times 10^{-5}$ & $2\times 10^{-5}$ & $2\times 10^{-5}$ \\
Feature-head LR & -- & -- & $10^{-3}$ & $10^{-3}$ \\
Weight decay & 0.01 & 0.01 & 0.01 & 0.01 \\
Warmup ratio & 0.1 & 0.1 & 0.1 & 0.1 \\
LR scheduler & linear warmup & linear warmup & linear warmup & linear warmup \\
Optimizer & AdamW & AdamW & AdamW & AdamW \\
Gradient clipping & 1.0 & 1.0 & 1.0 & 1.0 \\
Dropout & 0.1 & 0.1 & 0.1 & 0.1 \\
Loss & CE & CE & CE & CE \\
\bottomrule
\end{tabular}%
}
\end{table*}

\begin{table*}[t]
\centering
\caption{Handcrafted feature categories ($m{=}62$ total; $k{=}30$ selected).}
\label{tab:features_categories}
\small
\setlength{\tabcolsep}{6pt}
\renewcommand{\arraystretch}{1.05}
\begin{tabular}{@{}lcl@{}}
\toprule
\textbf{Category} & \textbf{Count} & \textbf{Description (examples)} \\
\midrule
Perplexity & 8 & GPT-2 perplexity statistics (doc/token) \\
Entropy & 10 & character/word/ngram/POS entropy \\
Burstiness & 6 & repetition/interval burstiness measures \\
Repetition & 10 & self-BLEU, unique n-gram ratios, compression, dup.\ sentences \\
Vocabulary & 8 & Zipf/Yule/Hapax/TTR/MATTR, lexical density \\
Coherence & 6 & sentence similarity, topic drift, consistency \\
Readability & 6 & Flesch, FK-grade, ARI, Coleman--Liau, SMOG, Dale--Chall \\
Stylometric & 8 & function-word ratio, pronouns, punctuation density, caps ratio \\
\midrule
\textbf{Total} & \textbf{62} & \\
\textbf{Selected} & \textbf{30} & via combined importance scoring \\
\bottomrule
\end{tabular}
\end{table*}

\subsubsection{Complete Feature List (62)}
\label{app:features_list}

\noindent\textbf{Perplexity (8):}
\texttt{ppl\_text, ppl\_mean\_token, ppl\_max\_token, ppl\_std\_token, ppl\_skew\_token, ppl\_low\_ratio, ppl\_high\_ratio, ppl\_trend.}

\noindent\textbf{Entropy (10):}
\texttt{ent\_char, ent\_word, ent\_bigram, ent\_trigram, ent\_conditional, ent\_rate, ent\_pos, ent\_punct, ent\_sent\_len, ent\_word\_len.}

\noindent\textbf{Burstiness (6):}
\texttt{burst\_word, burst\_sent\_len, burst\_punct, burst\_memory, burst\_fano, burst\_iat\_var.}

\noindent\textbf{Repetition (10):}
\texttt{rep\_self\_bleu2, rep\_self\_bleu3, rep\_self\_bleu4, rep\_unique\_bigram, rep\_unique\_trigram, rep\_bigram\_rate, rep\_trigram\_rate, rep\_max\_ngram\_freq, rep\_compression, rep\_dup\_sent.}

\noindent\textbf{Vocabulary (8):}
\texttt{vocab\_zipf\_coef, vocab\_yules\_k, vocab\_hapax, vocab\_dis, vocab\_heaps, vocab\_ttr, vocab\_mattr, vocab\_lexical\_density.}

\noindent\textbf{Coherence (6):}
\texttt{coh\_sent\_sim\_mean, coh\_sent\_sim\_var, coh\_topic\_drift, coh\_self\_sim, coh\_semantic\_density, coh\_consistency.}

\noindent\textbf{Readability (6):}
\texttt{read\_flesch, read\_fk\_grade, read\_ari, read\_coleman, read\_smog, read\_dale\_chall.}

\noindent\textbf{Stylometric (8):}
\texttt{style\_func\_word, style\_pronoun, style\_conjunction, style\_avg\_word\_len, style\_avg\_sent\_len, style\_sent\_len\_var, style\_punct\_ratio, style\_cap\_ratio.}

\subsubsection{Feature Normalisation and NaN/Inf Handling}
We normalise features using \texttt{RobustScaler} (median/IQR) fitted on the training set and applied to validation/test sets, and replace NaN/Inf values with zeros after extraction.

\subsubsection{Global Feature Selection ($k{=}30$ from $m{=}62$)}
\label{app:feature_selection}

Feature selection is performed \emph{once} using source-side training data only (HC3~PLUS \texttt{train}) to avoid target-domain leakage. We compute a global importance score per feature by combining (i) mutual information with the binary label \citep{CoverThomas06-InfoTheory} and (ii) the absolute point-biserial correlation \citep{Tate54-PointBiserial}. After max-normalising both scores to $[0,1]$, we average them and select the top-$k$ features under the resulting ranking.

\subsection{Feature-Attention Module and Classifier Head}
\label{app:feat_attention}

\paragraph{Architecture.}
Given selected features $f_k(x)\in\mathbb{R}^{30}$, the feature-attention module produces attention weights $a(x)\in\mathbb{R}^{30}$ via an importance MLP followed by softmax; weighted features are projected to $z_f(x)\in\mathbb{R}^{128}$. The transformer pooled vector $h_{\texttt{[CLS]}}\in\mathbb{R}^{768}$ is concatenated with $z_f(x)$ to form a 896-dim fused vector, which is classified by an MLP head.

\begin{table}[t]
\centering
\caption{Feature-attention module and classifier architecture.}
\label{tab:feat_attention_arch}
\small
\setlength{\tabcolsep}{6pt}
\renewcommand{\arraystretch}{1.05}
\begin{tabular}{@{}llcc@{}}
\toprule
\textbf{Component} & \textbf{Layer} & \textbf{In} & \textbf{Out} \\
\midrule
\multirow{4}{*}{Importance Net}
& Linear + LayerNorm & 30 & 128 \\
& ReLU + Dropout(0.1) & 128 & 128 \\
& Linear & 128 & 30 \\
& Softmax & 30 & 30 \\
\midrule
\multirow{3}{*}{Output Proj}
& Linear + LayerNorm & 30 & 128 \\
& ReLU + Dropout(0.1) & 128 & 128 \\
& Linear & 128 & 128 \\
\midrule
\multirow{4}{*}{Classifier}
& Dropout(0.1) & 896 & 896 \\
& Linear + LayerNorm & 896 & 384 \\
& ReLU + Dropout(0.1) & 384 & 384 \\
& Linear & 384 & 2 \\
\bottomrule
\end{tabular}
\end{table}

\subsection{Data and Preprocessing Notes}
\label{app:preproc}

\paragraph{Filtering.}
We discard extremely short texts (minimum length $\ge$ 50 characters and minimum words $\ge$ 10) to ensure stable feature extraction.

\paragraph{Cleaning.}
We apply minimal cleaning to preserve stylometric signals: no global lowercasing and no removal of punctuation/stopwords; we apply Unicode normalisation and whitespace normalisation; HTML tags may be removed if present.

\paragraph{Tokenization.}
Transformer tokenization uses the HuggingFace tokenizer for each backbone with max length 512, truncation enabled, and padding to \texttt{max\_length} \citep{Wolf20-Transformers}.

\subsection{Evaluation on \dataset{AI-Text-Detection-Pile}}
\label{app:pile_full}
We evaluate on the \textbf{full} \dataset{AI-Text-Detection-Pile} test distribution (no balanced subsampling), and report balanced accuracy and class-wise recall due to class imbalance.

\end{document}